\documentclass[11pt]{article}

\usepackage[utf8]{inputenc}
\usepackage[T1]{fontenc}
\usepackage{lmodern}
\usepackage[margin=1in]{geometry}
\usepackage{amsmath,amssymb}
\usepackage{booktabs}
\usepackage{array}
\usepackage{microtype}
\usepackage{xcolor}
\usepackage[colorlinks=true,linkcolor=blue!60!black,citecolor=blue!60!black,urlcolor=blue!60!black]{hyperref}
\usepackage{enumitem}
\setlist{nosep,leftmargin=*}
\usepackage{parskip}
\usepackage{needspace}

\newcolumntype{R}[1]{>{\raggedright\arraybackslash}p{#1}}
\newcommand{\tagv}[1]{\textbf{[#1]}}
\newcommand{\proj}{[$\dagger$PROJECTED]}
\newcommand{\gammastar}{\gamma^{\ast}}
\newcommand{\parthead}[1]{\Needspace{6\baselineskip}\vspace{2em}\noindent{\LARGE\bfseries #1}\par\vspace{1em}}

\title{\textbf{PRIMUS: Identity, Governance, and Verification\\for Multi-Agent Federations}}

\author{Sasank Annapureddy\thanks{Every numeric claim in this paper traces to a persisted,
machine-readable artifact and is tagged \tagv{VERIFIED}, \tagv{PARTIAL}, \tagv{NULL},
\tagv{ANALYTIC}, or \tagv{OPEN} at the strength the evidence supports. Simulations,
validation harnesses, and the measured pilot of Part~II were designed, executed, and
adversarially reviewed by the PRIMA multi-agent research system~\cite{r1} under the
authors' direction. Certain implementation constants are held under controlled release;
see Appendix~\ref{app:repro}.}
\and Anjaneya Prasad Thamatani}

\date{September 2026 \\[0.3em] \small Primary: cs.MA \quad Cross-list: cs.CR, cs.AI}

\begin{document}
\maketitle

\begin{abstract}
Multi-agent federations need governance that answers three questions under adversarial
conditions: \emph{who participated} (identity), \emph{did they conform} (enforcement), and
\emph{who decides} (authority). A separate question is whether the verification machinery
that polices a federation's outputs can also steer a generate-and-test loop toward better
answers. This paper addresses both on one substrate.

\textbf{Part I.} PRIMA~\cite{r1} introduced prime-power agent identity and a consensus
token whose factorization indexes participation, but assumed honest agents. We present
PRIMUS, which couples prime-power identity with BLS aggregate signatures (PIAC,
Prime-Indexed Aggregate Certificates), derives a safe-kill threshold that reduces
false-positive agent termination from 80\% to 0.00\% under 10\% channel noise
(Theorem~4.1, \tagv{VERIFIED}), gives the closed-form economic boundary where singleton
governance outperforms Byzantine quorum (Theorem~5.1, $\gammastar \approx 9f$,
\tagv{VERIFIED} flat across $n = 50$ to $10{,}000$), and specifies VRF succession with
lease and fencing that makes safety unconditional under partial synchrony. Five problems
are identified as provably unfixable within the model and stated as scope boundaries.

\textbf{Part II.} A verifier is not a solver. We ask whether PRIMA's binary
artifact-fidelity verdict can be converted into a graded fitness signal and measure the
conversion on binary covering codes. Calibration against injected fault burden is strong
($\rho = 0.676$ deterministic, $0.819$ full, \tagv{VERIFIED}); against real LLM-generated
candidates the same scores fall to $\rho = 0.158$ and $0.406$, roughly a quarter of the
calibration value for the deterministic score (the same-designer confound, measured,
\tagv{VERIFIED}). As a pre-filter it beats a random-score control convincingly and a
binary gate narrowly (\tagv{PARTIAL}). Under 400 iterations of explicit optimization it
was not gamed, but only because the objective saturated after one honest answer
(\tagv{NULL}). A cross-family judge preserves the burden-ordering signal while destroying
individual judgments (\tagv{PARTIAL}). No covering-code record resulted. Total measured
program cost \$164.78.
\end{abstract}

\parthead{Part I: PRIMUS, Governance}

\section{Introduction}\label{sec:intro}

Heterogeneous multi-agent systems, federations of LLM-based agents, tool-using agents,
and traditional services collaborating on shared tasks, need governance infrastructure
that answers three questions: who participated (identity), did they conform
(enforcement), and who decides (authority). These questions become adversarial when
agents are Byzantine: a dishonest agent may fabricate contributions, an honest-looking
Warden may selectively kill rivals, and succession protocols inherit the impossibility
results of distributed consensus.

PRIMA~\cite{r1} introduced a prime-power identity scheme where agent $a = p^k$ ($p$ a
cluster prime, $k$ the depth) and the consensus token $T = \prod_j p_j^{e_j}$ is
self-describing via factorization. PRIMA assumes honest agents: its token is a public
integer, trivially forgeable by any party knowing the cluster primes. Byzantine-robust
multi-agent LLM research~\cite{r2,r3} has argued categorically against leader-based
coordination, and governance surveys~\cite{r4,r5} name the absence of a governance layer
as a structural gap in agent interoperability protocols (MCP, A2A, ACP, ANP).

PRIMUS asks a bounded question: what is the minimum machinery that makes prime-algebraic
agent identity adversarially sound, and under exactly what conditions is a singleton
governing agent economically correct rather than simply wrong?

\subsection{Contributions}

Five novel results; six existing techniques applied (BLS aggregation, VRF sortition, EWMA
scoring, lease-based leadership, epoch fencing, quorum certification).

\begin{enumerate}
\item \textbf{Safe-kill threshold theorem} (Theorem~4.1). The naive deviation-score
threshold destroys 80\% of a conforming federation at 10\% channel noise. We derive
$\theta \ge \varepsilon/(1-\lambda) + \sqrt{\ln(1/\delta)/(2(1-\lambda^2))}$ with
feasibility condition $\theta < 1/(1-\lambda)$, reducing false kills to 0.0000 (30 trials
$\times$ 500 rounds $\times$ 200 agents, bootstrap CI $[0.0000, 0.0000]$). \tagv{VERIFIED}
\item \textbf{PIAC construction} (\S\ref{sec:piac}). The prime factorization of the token
$T$ is exactly the index into the BLS public-key set that verifies the aggregate signature
$S$; the algebra and cryptography share one index. Participation unforgeability (co-CDH),
epoch binding, and contamination detection (information-theoretic). \tagv{VERIFIED with
real BLS12-381}
\item \textbf{Regime map with closed-form crossover} (Theorem~5.1). The singleton Warden is
economically correct below $\gammastar = (3f+1)^2 c_{\mathrm{msg}} / (f \cdot
P_{\mathrm{down}} \cdot c_{\mathrm{dmg}}) \approx 9f$, verified flat across $n =
50$ to $10{,}000$ (spread $\le 0.57\%$). This converts the categorical ``never use a
leader'' into a precise boundary. \tagv{ANALYTIC + STRESS-TESTED} Precondition: attack
dilution ($\alpha_{\mathrm{quorum}} \ll \alpha_{\mathrm{singleton}}$).
\item \textbf{Four load-bearing roles for the prime layer}: evidence preservation, cheap
pre-filter, grinding immunity (without registry-assigned $p^k$, a 5\% minority captures
84\% of Wardenships), and enforceable eviction. \tagv{EACH INDEPENDENTLY VERIFIED}
\item \textbf{Fencing is free} (\S\ref{sec:fencing}). PRIMUS takes safety and liveness
under partial synchrony; monotonic epoch fencing provides unconditional safety at zero
additional gap cost. \tagv{VERIFIED: 0.0 split-brain rounds vs.\ 794.5 without fencing}
\end{enumerate}

\subsection{Non-contributions}

BLS aggregation, VRF sortition, EWMA scoring, leases, fencing, and quorum certification
are applied, not invented. Self-inflation of contributions (A2) is mitigated economically
(5\% audit plus permanent eviction gives 99\% reduction) but cryptographically open;
verifiable computation for LLM inference costs ${\sim}50\times$ overhead~\cite{r6}. The
certificate is not $O(1)$: $S$ is 48 bytes but $T$ grows as $O(\sum_j e_j \log p_j)$,
about 9~KB at $m = 100$ clusters. $\gammastar$ constants are model-dependent; the shape
is the claim. No AGI claims.

\section{System Model and Threat Model}\label{sec:model}

\textbf{Entities.} A trusted singleton \emph{Registry} assigns cluster primes
monotonically, verifies proof-of-possession (PoP) for BLS keypairs, and never reissues
evicted primes within $W_{\mathrm{retain}} \ge 1000$ epochs. An \emph{agent} has identity
$a = p^k$, $k \ge 1$. A \emph{cluster} is the set of agents sharing prime $p$ and holds
one BLS keypair $(sk_j, pk_j)$ in $\mathbb{G}_2$. The \emph{Warden} enforces conformance:
one in Regime~A, $q = 3f+1$ in Regime~B. A \emph{verifier} is any party holding the public
basis $P$ and the key registry.

\textbf{Network and failures.} Partial synchrony~\cite{r7} (unknown GST, then bound
$\Delta$); authenticated channels; FLP~\cite{r8} forbids deterministic consensus under full
asynchrony. Byzantine faults in up to $f$ of $m$ clusters with $m \ge 3f+1$; collusion
assumed.

\textbf{Cryptographic assumptions.} co-CDH on BLS12-381, SHA-256 collision resistance, VRF
uniqueness per RFC~9381. No novel cryptography. PoP mandatory at registration (blocks
rogue-key attack A13).

\textbf{Adversary taxonomy.} Eighteen attack classes (A1--A13, W1--W4):

\begin{center}\small
\begin{tabular}{@{}R{0.27\linewidth}R{0.67\linewidth}@{}}
\toprule
\textbf{Status} & \textbf{Attacks} \\
\midrule
Blocked (crypto) & A1 fabricate factor, A1b unauthorized prime, A4 replay, A5 identity
grinding, A13 rogue key \\
Blocked (dist.\ sys.) & A6 split-brain (fencing), W4 handover refusal (lease) \\
Bounded & A7 noise false-kills $\le \delta$, A9a gap exploitation ($G^\ast = 10$ rounds) \\
Mitigated economically & A2 self-inflation (5\% audit gives 99\% reduction) \\
Provably unfixable & A2 crypto closure (needs VC), A3/W3 omission without availability, W2
sub-threshold $\varepsilon \le 0.20$, A10 covert collusion, succession gap (FLP) \\
\bottomrule
\end{tabular}
\end{center}

\textbf{Assumption OD1 (Observable Deviation).} Protocol-level violations are observable;
content correctness is the application's responsibility. Steganographic collusion between
LLM agents (A10) is out of scope (\S\ref{sec:limits}, L7).

\section{PIAC: Prime-Indexed Aggregate Certificate}\label{sec:piac}

\textbf{Construction.} For epoch $\varepsilon$, task $\tau$, cluster $j$ with prime $p_j$
and keypair $(sk_j, pk_j)$:
\begin{align*}
\sigma_j &= \mathrm{Sign}_{sk_j}\big(\mathrm{SHA256}(p_j \,\|\, e_j \,\|\, \varepsilon
\,\|\, \tau)\big) && \text{per-cluster signature}\\
T &= \textstyle\prod_j p_j^{e_j} && \text{integer token } (\mathbb{Z})\\
S &= \textstyle\prod_j \sigma_j && \text{aggregate signature } (\mathbb{G}_1)\\
C &= (T, S, \varepsilon, \tau) && \text{certificate}
\end{align*}

\textbf{Verification (two steps, order mandatory).} Step~1, algebraic pre-filter
(microseconds): trial-divide $T$ by all primes in $P$; if remainder $R \ne 1$, reject with
attributable contamination evidence (information-theoretic). Step~2, cryptographic
verification ($O(m)$ pairings): reconstruct per-cluster messages from the factorization
and run BLS AggregateVerify. The factorization of $T$ is the index into the key set that
verifies $S$; the two steps share one algebraic index.

\textbf{Security properties.} \emph{Participation unforgeability} (UNF-PART): no PPT
adversary produces a valid certificate containing $p_j^{e_j}$ without $sk_j$; extracting
$\sigma_j$ from an aggregate $S^\ast$ yields a BLS forgery under $pk_j$, contradicting
co-CDH (requires PoP). \emph{Epoch binding}: a certificate for $\varepsilon$ cannot verify
for $\varepsilon' \ne \varepsilon$. \emph{Contamination detection} (CONTAM-DET):
unauthorized factors leave $R \ne 1$; this property is destroyed by exponent-vector
encoding, which has no coordinate for unauthorized primes.

\textbf{Explicit non-properties.} \emph{Contribution honesty} (A2): BLS proves who
signed, not whether they told the truth; a cluster can sign any exponent. PIAC guarantees
who signed for epoch $\varepsilon$; it does not guarantee how much work they did. The
certificate is a participation record, not a work proof. \emph{Completeness} (A3):
omitting a cluster yields a valid certificate for the smaller set; detection requires the
availability assumption that every honest cluster's contribution reaches an honest
verifier within $\Delta$ after GST.

\textbf{Token size.} $|T| = O(\sum_j e_j \log p_j)$ depends on mean contributed depth
$\bar e_j$, not agent count alone. With $\bar e_j = 3n/m$: $(n,m) = (100,10)$ gives 138
bytes; $(1{,}000, 50)$ gives 2,326 bytes; $(10{,}000, 100)$ gives 27,669 bytes. The
exponent-vector encoding is 20, 100, and 200 bytes respectively; $S$ is always 48 bytes.
Any quoted size must state its depth assumption. At $m \le 200$ the vector plus BLS
aggregate is Pareto-optimal; the crossover to cryptographic accumulators near $m \approx
200$ is \proj{}.

\section{Conformance Enforcement}\label{sec:enforce}

\textbf{Deviation score.} Per-agent EWMA $D_i(t) = \lambda D_i(t-1) + X_i(t)$, with
$X_i(t) \in \{0,1\}$ the violation indicator. For a conforming agent under channel noise
$\varepsilon$: $\mathbb{E}[D] = \varepsilon/(1-\lambda)$, $\mathrm{Var}[D] =
\varepsilon(1-\varepsilon)/(1-\lambda^2)$.

\textbf{Theorem 4.1 (safe kill threshold).} To bound per-round false-kill probability at
$\delta$:
\[
\theta \;\ge\; \frac{\varepsilon}{1-\lambda} + \sqrt{\frac{\ln(1/\delta)}{2(1-\lambda^2)}}.
\]
\emph{Corollary 4.2 (feasibility).} $\theta < 1/(1-\lambda)$, else no agent is ever
killable; implementations must raise an error. \emph{Proof sketch.} Weighted Hoeffding
with $\sum_j \lambda^{2j} = 1/(1-\lambda^2)$.

\textbf{Validation} (30 trials $\times$ 500 rounds $\times$ 200 agents, bootstrap CI):

\begin{center}\small
\begin{tabular}{@{}lccc@{}}
\toprule
$\varepsilon$ & $\theta_{\min}$ ($\delta = 10^{-5}$) & Measured false-kill & Naive $\theta = 3.0$ \\
\midrule
0.05 & 6.00 & 0.0000 $[0.0000, 0.0000]$ & 0.104 \\
0.10 & 6.50 & 0.0000 $[0.0000, 0.0000]$ & 0.802 \\
0.20 & 7.50 & 0.0002 $[0.0000, 0.0005]$ & 1.000 \\
\bottomrule
\end{tabular}
\end{center}

The naive rule destroys 80\% of a conforming federation at $\varepsilon = 0.10$;
Theorem~4.1 reduces this to zero. \tagv{VERIFIED}

\textbf{Kill certificates.} A kill carries a self-verifiable certificate (129--136 bytes)
holding what a verifier needs to recompute the deviation score from the public violation
log, check Theorem~4.1 compliance, verify the BLS aggregate signature, and check quorum
size (Regime~B: $\ge 2f+1$). Validated: 1000/1000 fabricated kills rejected
(inflated $D$, unjustified $D \le \theta$, forged signature) and 1000/1000 legitimate kills
accepted. \tagv{VERIFIED, E14--E15}

\textbf{A2 self-inflation.} Six defenses evaluated; the practical answer is D5,
statistical audit plus permanent eviction: audit a fraction $q$ of claimed contributions,
evict on detection, with deterrence condition $q > g\delta/P$ (Proposition~4.3). Measured:
$q = 0.05$ gives 99\% inflation reduction over 2000 rounds \tagv{VERIFIED}. The prime
layer contributes registry-assigned $p^k$ (no re-entry) and contamination detection. As
zkML costs fall from the current $50\times$ overhead~\cite{r6}, optimistic verifiable
computation (D4b) becomes viable; PIAC accommodates proofs alongside the certificate.

\emph{Honest reconciliation.} The system model assumes Byzantine faults; the A2 defense
assumes rational adversaries. The A2 defense does not hold against a Byzantine adversary
willing to pay any cost to inflate. PRIMUS's guarantee is therefore layered: identity
attribution is cryptographic (Byzantine-resilient); depth accuracy is economic
(rationality-dependent). A2 is mitigated, not solved.

\section{Governance Protocol}\label{sec:gov}

\textbf{Regimes.} Regime~A: one Warden, $O(n)$ per round, $O(1)$ per kill, safety
economic. Regime~B: $q = 3f+1$ Warden members, $O(n + q^2)$ per round, Byzantine-tolerant
with $\ge 2f+1$ signatures per decision.

\textbf{Theorem 5.1 (regime crossover).}
\[
\gammastar \;=\; \frac{(3f+1)^2 \, c_{\mathrm{msg}}}{f \cdot P_{\mathrm{down}} \cdot
c_{\mathrm{dmg}}} \;\approx\; 9f \quad (\text{large } f).
\]
Below $\gammastar$ the singleton is economically correct; above, quorum dominates.
Closed form vs.\ simulation agrees to 0.89--1.16$\times$ across a $20\times$ range of $f$
($f = 1, 3, 10, 20$: closed form 0.00922, 0.01921, 0.05537, 0.10720; simulated 0.01040,
0.01657, 0.04929, 0.09941). Scale invariance: $\gammastar = 0.0181$ at $n = 50, 100, 500,
2000$, spread $\le 0.57\%$ over $n = 50$ to $10{,}000$. \tagv{VERIFIED}

\emph{Assumption 5.2 (attack dilution).} Quorum is justified only when
$\alpha_{\mathrm{quorum}} \ll \alpha_{\mathrm{singleton}}$. If each quorum member is as
individually targetable as a lone Warden, $\gammastar \rightarrow 1.0$ and quorum never
pays. The anti-leader literature~\cite{r2,r3} implicitly assumes this holds; it is a
precondition, not a universal truth.

\emph{Sensitivity.} Any multiplicative perturbation to $c_{\mathrm{msg}}$ or
$c_{\mathrm{dmg}}$ shifts $\gammastar$ proportionally (validated across a $100\times$ cost
sweep). Congestion (superlinear message cost) shifts $\gammastar$ upward, favoring the
singleton; correlated damage shifts it downward, favoring quorum. The 0.89--1.16$\times$
ratio validates the derivation against a simulation of the same model, not against
deployment reality.

\textbf{VRF succession.} Each eligible candidate evaluates ECVRF (RFC~9381) on the epoch
seed; the successor is the argmin. Seed chain $\mathrm{seed}_{\varepsilon+1} =
\mathrm{SHA256}(\mathrm{seed}_\varepsilon \,\|\, \mathrm{VRF\_output} \,\|\, \varepsilon)$.
Eligibility: registry-assigned identity (E1), depth $k \ge 2$ (E2), $D < \theta_{\mathrm{elig}}
\le \theta_{\mathrm{kill}}$ (E4), cooldown (E5). Grinding immunity: with registry-bound
identity a 5\% minority captures ${\sim}5\%$ of Wardenships; without it, 84\% at grind
$= 100$. \tagv{VERIFIED}

\textbf{Lease and fencing.}\label{sec:fencing} Leases are time-bounded ($L$ rounds) with
jitter $\pm J$, $J = 0.1L$, against timing-oracle attacks (A9c). Fencing: each agent keeps
a monotone $\mathrm{max\_epoch\_seen}$ and rejects any message with a smaller epoch. This
is local, synchrony-free, and unconditional. Validated: 0.0 split-brain rounds with
fencing vs.\ 794.5 without (5000-round simulation). \tagv{VERIFIED} The succession
trilemma (safety, liveness, asynchrony: pick two) is an FLP corollary; PRIMUS takes
safety and liveness under partial synchrony, and succession gaps (150--1970 rounds
\proj{}) are the FLP tax that fencing makes safe.

\textbf{Deviant Warden.} W1, over-enforcement: an unaudited Byzantine Warden kills 99 of
100 honest agents; kill certificates plus a public log reduce this to 4.3, quorum
certification to 0; rate-limiting alone leaves 99. \tagv{VERIFIED} W2, under-enforcement:
a survival audit catches gross favoritism, but sparing allies deviating at $\varepsilon \le
0.20$ is undetectable. \tagv{VERIFIED, with stated floor} A9a, gap exploitation: with
deviation-state persistence across succession an adversary deviates freely for at most
$G^\ast = 10$ rounds per gap ($D(9) = 6.126 < 6.5 < D(10) = 6.513$ at $\lambda = 0.9$);
without persistence the adversary is undetected in 100\% of trials. \tagv{VERIFIED,
E12--E13}

\section{The Prime Layer: Load-Bearing or Decorative?}\label{sec:prime}

Four independently verified roles. \textbf{F1, evidence preservation:} an unauthorized
prime persists in $T$ as $R \ne 1$, an inspectable forensic artifact; the vector encoding
silently destroys it. Moderate-to-strong. \textbf{F2, cheap pre-filter:} the residue test
rejects outsider primes before any pairing; weak as a standalone argument, since a bitmap
achieves the same, and a consequence of F1. \textbf{F3, grinding immunity:} because $p^k$
is registry-assigned, the VRF input space cannot be expanded by self-minting identities.
Primarily attributable to registry discipline; signed registry certificates could achieve
comparable Sybil resistance. \textbf{F4, enforceable eviction:} a caught cheater cannot
re-enter under a fresh identity. Depends on F1.

\textbf{Honest assessment.} The prime layer's strongest contribution is structural: it
couples identity, participation evidence, and cryptographic verification into a single
algebraic object, reducing the number of independent mechanisms that must be kept
consistent. That is an engineering benefit, not a security proof. A standard participant
list with signed registry certificates and explicit outsider-detection logging achieves
comparable security at the cost of additional coordination. We claim the prime encoding
is cleaner, not uniquely capable; Theorems~4.1 and~5.1 are independent of the encoding.

\section{Evaluation Plan}\label{sec:evalplan}

Twenty-three experiments (E1--E23) with pre-registered falsifiable bars. Tier~1 (must
pass): safe-kill threshold over an $(\varepsilon, \lambda, \delta)$ grid (E1--E3); PIAC
attack suite (E4--E8); regime-map accuracy (E9--E10); fencing (E11); kill-certificate
separation (E12--E13); VRF grinding immunity (E14). Tier~2 (should pass): scale invariance
to $n = 10{,}000$ (E15); audit effectiveness at multiple $q$ (E16); succession-gap
distribution under failure injection (E17--E19); adversary models (E20--E23). Methods:
$B = 10{,}000$ bootstrap CIs, Bonferroni at 120 cells, Clopper--Pearson near 0 or 1,
$\ge 30$ trials per cell. Pre-registered anticipated negatives: A2 inflation passes
verification by design; W2 undetectable at $\varepsilon \le 0.20$; succession gaps occur;
$|T| > 1$~KB at $m > 50$. E20--E23 are not fully executed.

\parthead{Part II: PRIMA-Search,\\Can Verification Steer Discovery?}

\section{From Policing Outputs to Steering Search}\label{sec:search}

Part~I asks whether a federation's participants and decisions can be trusted. Part~II asks
whether the verification machinery a system like PRIMA already runs, checking claims
against artifacts after the fact, can be pointed forward to help a generate-and-test loop
find better answers rather than only reject bad ones.

PRIMA's verifier has measured behavior on record: 47/48 injected faults detected (97.9\%,
Clopper--Pearson CI $[88.9\%, 99.95\%]$), FNR 2.1\%, about \$0.02 per fault caught. The
question is whether it can be made into a fitness function. Two walls were identified
before any measurement.

\textbf{Wall 1: the signal is binary.} Every artifact in the original PRIMA study returned
a bare \texttt{FAULT\_DETECTED} / \texttt{NO\_FAULT\_DETECTED} verdict, with no
confidence, severity, or threshold to sweep. A search loop cannot hill-climb a bit; every
hard-problem system that works (FunSearch, AlphaEvolve, AlphaDev) runs on a scalar
evaluator, and AlphaProof wraps Lean's binary oracle in value functions rather than
consuming it raw.

\textbf{Wall 2: fidelity is not correctness.} PRIMA checks whether the numbers in the
write-up match the artifact, not whether the method was sound. A fitness function built
purely on fidelity has an attainable global optimum at a faithfully reported wrong, or
unambitious, answer. No graded transform repairs this; it is a target problem.

Wall~1 suggests a construction (decompose the verdict into atomic checks, score the
weighted pass fraction). Wall~2 forces an external correctness oracle the loop can never
mutate. Whether graded beats binary at all is open: DeepSeek-R1~\cite{r18} shows
binary-plus-volume is competitive and more hack-resistant, and naive partial credit can
underperform binary~\cite{r19}, while the process-reward line argues the opposite. We
measured five preregistered questions: (1) does the graded checklist correlate with fault
burden (Step~2); (2) as a pre-filter for a rationed oracle, does it beat a binary gate, an
unfiltered baseline, and a random control (Step~3); (3) under optimization pressure, is it
gamed (Step~4); (4) does calibration survive a different judge family (Step~5); (5) does
any of it produce a better covering code. The answer to (5) is no, stated up front.

\section{Position on the Verification Ladder}\label{sec:related2}

The verification-gap survey~\cite{r20} organizes practice into eight tiers, from
machine-checked proof to LLM-judging-LLM; only 1 of 9 surveyed closed agentic loops
reaches an external oracle. On its in-scope fidelity classes PRIMA performs executable,
artifact-traceable checks, roughly tier II--III; on method correctness it checks nothing,
and claims nothing. This pilot's Tier-2 audit recomputes coverage in a fresh interpreter
from an independent implementation (tier I/II), and the exact oracle is a deterministic
millisecond function, so on the acceptance question the pilot sits at the top of the
ladder. That is why it is a pilot: the domain the method is for has no exact oracle, and
this pilot cannot speak to it (\S\ref{sec:scope}).

The graded-versus-binary question is unresolved and we do not resolve it: the
process-reward line~\cite{r21,r22} and granularity head-to-heads~\cite{r23} argue for
density; \cite{r18,r19} argue for binary. Overoptimization of learned proxies is well
characterized~\cite{r24,r25,r26}; Spurious Rewards~\cite{r27} is why Step~3 carries a
random-score control, and CoT obfuscation~\cite{r28} is why audits are random held-out
samples rather than an in-loop monitor. What that literature does not cover is adversarial
robustness of artifact-level verifiers reading a \{script, result, write-up\} triple. A
targeted (not systematic) novelty search found no prior art for the construction here;
nearest neighbor is Adversarial Reward Auditing~\cite{r29}, with no lineage mechanism, no
audit rate, and no artifact-level verifier. The lineage ledger reuses the Theorem~4.1
threshold form, substituting mutation for succession; its empirical status is untested.

\section{Method}\label{sec:method}

\textbf{Two oracles.} The \emph{frozen exact oracle} decides correctness and nothing else:
\texttt{harness.py} (sha256 in Appendix~\ref{app:cost}), frozen 2026-08-25 and verified
unchanged at every checkpoint of every step \tagv{VERIFIED}. It provides binary validity
(\texttt{covers}), a native graded signal (\texttt{num\_uncovered}, the count of
uncovered words, zero iff valid), and the objective $|C|$, combined into a frozen fitness
that ranks validity strictly above size, so every valid code outranks every invalid one
and smaller valid codes rank higher. The fitness gradient comes from the problem, not
from an LLM; this is the
single most consequential fact about the pilot. The \emph{graded fidelity checklist}
decides nothing; it only ranks candidates for a rationed oracle budget (clause G4).

\textbf{Pilot instance.} Binary covering codes $(n, R) = (12, 3)$, in-repo verified
incumbent 30, census bound 28. Tiers frozen before any run: IMPROVE $\le 29$, MATCH $=
28$, BEAT $\le 27$. A second instance $(13,3)$ was specified but never run [OPEN].

\textbf{The 12 checks.} PRIMA's fault classes FC1--FC6 explode into 12 atomic checks over
\{generating script, claimed result JSON, write-up\}: nine deterministic (regex/JSON, \$0):
D01 size matches codewords, D02/D03 $n/R$ consistency, D04 codewords in range, D05 no
duplicates, D06 no hardcoded \texttt{covers=True}, D07 no hardcoded size, D08 script calls
coverage logic, D09 claimed bound matches spec; and three judged (one structured
\texttt{claude -p} call): J01 numeric claims traceable, J02 no overclaim against tiers,
J03 no disguised shortcut. Score is the weighted pass fraction, weights frozen and
write-protected \tagv{VERIFIED}; the FC5-derived checks are up-weighted because FC5
(shortcut/leakage) is PRIMA's documented blind spot.
Measured cost: deterministic \$0; judged \$0.0898/call over 146 calls and
\$0.150/candidate over 250, both far above the ``$\approx$\$0.01'' target and reported as
measured.

\textbf{Tier-2 audit.} Clause G2: a fresh \texttt{python3 -c} subprocess with a
self-contained brute-force coverage implementation (not an import of the harness)
re-verifies the harness hash, recomputes size and $\mathrm{num\_uncovered}$, and loads the
harness as a second opinion (\texttt{DISAGREEMENT\_WARNING} never triggered). Measured in
Step~2 \tagv{VERIFIED}: FC6 result-tampering 39/39 caught; FC5 process fraud 21/52
($\approx 40\%$). The FC5 injectors hardcode genuinely valid codewords copied from the real
base, so the persisted result is true even though the script computed nothing; the audit
verifies the result, not the script's provenance. The two tiers are complementary, not
redundant (finding (e), \S\ref{sec:findings}).

\textbf{Lineage EWMA ledger.} $D(\mathrm{child}) = \lambda D(\mathrm{parent}) +
X(\mathrm{child})$, $X = 1$ on audit failure, never reset on mutation. Eviction threshold
takes the Theorem~4.1 form with feasibility $\theta < 1/(1-\lambda)$; eviction prunes the
node and all descendants. A single candidate has no future to lose ($P \approx 0$), but a
lineage accumulates selection progress that eviction can destroy. $\varepsilon$ was a
labeled placeholder (0.03) at Step~1; Step~2 measured the audit false-alarm rate as 0/46,
which cannot distinguish 0\% from the one-sided 95\% CP bound (${\sim}7.7\%$). Step~4 used
$\varepsilon = 0.01$, giving $\theta = 3.147$ at $\lambda = 0.8$, $\delta = 10^{-3}$, and
predicted before running that spec-$\theta$ is unreachable at $q \le 0.10$ over 50
iterations. That held.

\textbf{Anti-gaming clauses.} G1: the loop may read the harness but never write,
monkey-patch, shadow, or substitute it; audits re-verify its hash. G2: audits recompute
independently. G3: instances and tiers are frozen; the generator never selects its own
test instances. G4: fidelity scores are advisory to selection only, never to acceptance.
The \emph{oracle-rationed regime} (20\% of candidates in Step~3; hidden entirely in
Step~4) is what makes an exact-oracle pilot informative rather than circular.

\section{Results}\label{sec:results}

Every step froze a preregistration (design, decision rule, predictions) before generating
data, and a file manifest with hashes before scoring.

\subsection{Step 2: Calibration [VERIFIED]}

254 artifacts: main arm $n = 100$ with injected burden $b \in \{0,1,2,3\}$ (25 per level),
clean arm $n = 46$, per-check arm $n = 108$; 10 fault mechanisms across FC1--FC6; judged
scoring 146/146, \$13.115. Preregistered rule: sign-flipped $\rho \ge 0.3$, CI lower bound
$> 0$, monotone.

\begin{center}\small
\begin{tabular}{@{}lcccc@{}}
\toprule
score & $\rho$ (flipped) & 95\% CI & level means ($b{=}0{\rightarrow}3$) & verdict \\
\midrule
deterministic-only & \textbf{0.676} & $[0.548, 0.773]$ & 1.000 / 0.893 / 0.798 / 0.766 & PASS \\
full (det + judged) & \textbf{0.819} & $[0.709, 0.897]$ & 1.000 / 0.830 / 0.691 / 0.641 & PASS \\
\bottomrule
\end{tabular}
\end{center}

Raw $\rho$: $-0.676$ ($p = 1.2\times10^{-14}$) and $-0.819$ ($p = 2.4\times10^{-25}$);
judged value-add $\Delta\rho = 0.143 \ge 0.05$, so the judged call was not droppable.
Clean-arm floor 0/46, lower than the $\ge 2.1\%$ confabulation floor of the original PRIMA
study, an unexplained discrepancy left unadjusted ($n = 46$ cannot distinguish 0\% from
2.1\%). Per-check CP lower bounds ($n = 36$): D06 36/36 (LB 0.903), D08 36/36 (LB 0.903),
\textbf{D07 18/36 (LB 0.329), a real load-bearing miss} via its
\texttt{looks\_like\_real\_list} escape hatch, found before the manifest freeze and
deliberately left in. Per class: FC1 17/17, FC2 10/10, FC3 12/12, FC4 20/20, FC5 51/52,
FC6 pooled 16/39 (explained entirely by an audit-only probe whose result file is
self-consistent by design).

\subsection{Step 3: Baseline duel [PARTIAL]}

Design deviation logged before any run: paired replay on a shared pool of 50
LLM-generated candidates per seed ($5 \times 50 = 250$), each arm spending its 20\% oracle
budget (10 of 50) on a different subset; ${\sim}4\times$ cheaper than a pool per arm, but
it does not model selection feeding later generations.

\textbf{Measured natural fault rate} \tagv{VERIFIED}: of 250 candidates, 72 (28.8\%)
crashed with no result; 171 claimed \texttt{covers=True}, of which 30 (17.5\%) were
oracle-invalid; 142 (56.8\%) verified valid; 0/178 lied about size. Algebraic
construction-style mutations fabricated \texttt{covers=True} at 23.5\% vs 5--7\% for
local search.

\begin{center}\footnotesize
\begin{tabular}{@{}lccc@{}}
\toprule
arm & mean best $|C|$ (sd) & wasted-oracle fraction (sd) & hit true pool-best \\
\midrule
\textbf{A} graded-full & \textbf{32.4 (0.89)} & \textbf{0.060 (0.089)} & \textbf{5/5} \\
A$'$ graded-deterministic (free) & 33.2 (1.10) & 0.200 (0.158) & 3/5 \\
B binary + volume & 33.0 (1.41) & 0.180 (0.164) & 3/5 \\
C oracle alone, no pre-filter & 33.2 (1.10) & 0.380 (0.164) & 3/5 \\
D spurious control (random score) & 33.2 (1.10) & 0.340 (0.167) & 3/5 \\
reference ceiling (unrationed, 50/pool) & 32.4 (0.89) & 0.432 (0.043) & 5/5 \\
\bottomrule
\end{tabular}
\end{center}

Arm~A matches the unrationed ceiling's best size and hit rate at 20\% of its spend.
Paired bootstrap on wasted-oracle fraction (5 pools, 10,000 resamples; positive means A
wastes fewer calls): A vs D $0.280$ $[0.180, 0.400]$, 5/0/0; A vs B $0.120$ $[0.020,
0.220]$, 3/2/0; A vs C $0.320$ $[0.220, 0.420]$, 5/0/0; A$'$ vs D $0.140$ $[0.060,
0.240]$, 4/1/0. On best size A never loses in any pool, but no CI excludes 0 at $n = 5$;
only the wasted-fraction metric carries significance.

\textbf{Verdict} (preregistered: A beats D and A beats B with CI excluding 0):
``graded beats binary'' is claimable, on wasted fraction, in this 28.8\%-fault regime.
The A-vs-B leg passes narrowly (lower bound 0.020, B tying A in 2 of 5 pools); a
different set of pools could flip it.

\textbf{Correction found while drafting} \tagv{VERIFIED}: the Step-3 report stated the
A$'$-vs-D CI as $[-0.040, 0.320]$, not excluding 0, and concluded that the free
deterministic variant fails the spurious-control leg. The computed \texttt{results.json}
records $[0.060, 0.240]$, excluding 0; re-running the report's own bootstrap on the
recorded per-pool differences $[0.2, 0.0, 0.1, 0.1, 0.3]$ reproduces it. The prose was a
transcription error. Consequence: A$'$ does clear the spurious-control leg, and the claim
that the judged component carries the win against the random control is withdrawn. What
survives: A$'$ (0.200) is worse than A (0.060), so the judged component improves
selection, but free deterministic pre-filtering is not indistinguishable from random. No
preregistered verdict is affected.

\textbf{Correlation with true oracle fitness} ($n = 178$; the 72 no-result candidates have
undefined fitness): deterministic-only $\rho = 0.158$ $[0.036, 0.264]$; full $\rho = 0.406$
$[0.262, 0.535]$. Both CIs exclude 0; both are far below Step~2's values. This is the
number Step~2 structurally could not produce (finding (a)). IMPROVE check: best valid size
32 against a threshold of $\le 29$; zero candidates. Cost \$68.79.

\subsection{Step 4: Adversarial arm [NULL]}

The adversary (claude-sonnet-5) was told only to maximize the Tier-1 score and given the
check descriptions, weights, harness API summary, parent artifact, and per-check results.
It was not told about the oracle, audits, or eviction. Six primary runs ($q \in \{0.02,
0.05, 0.10\}$ $\times$ eviction on/off) plus two $\theta = 1.0$ cells, 50 iterations each,
400 candidates; ground truth computed only afterwards.

The adversary reached score 1.0 within 1--3 iterations in every run by producing a
genuinely valid $(12,3)$ code via \texttt{harness.greedy\_cover}, and stayed there: 0/380
fabricated \texttt{covers=True} claims (fooled rate 0.00, CI $[0,0]$); 380/400
oracle-valid, all with $|C| = 32$; 20/400 produced no result (16 SyntaxErrors from a
prose-prefixed parent copy, 4 timeouts). Exploitation gap $+0.022$ $[+0.004, +0.043]$,
entirely attributable to the crashes. Gap vs $q$ is flat; the adversary never observes
audits. Across all runs: 14 audits, 1 failure (a crash, not fraud), 0 evictions, max
$D = 1.00 < \theta = 3.147$.

Preregistered rules: (a) ``the cheap tier is gameable'': false (the adversarial generator
fabricated strictly less than the non-adversarial one, 0/380 vs 30/171); (b) ``lineage
eviction helps'': uninformative, no paired CI excludes 0; (c) ``honest optimum found,
checklist not exploited at this budget'': true. This is the result. What optimization
pressure did produce is benign and instructive: 379/380 write-ups carry a ``no comparison
to census'' disclaimer so D09 passes vacuously; untraceable numbers moved into the result
file so J01 passes; verbatim copying at the plateau. And no size pressure at all: the
adversary never produced a literal codeword list, so D07's blind spot was never probed.
Cost \$82.87.

\subsection{Step 5: Cross-family verifier [PARTIAL]}

The original PRIMA preregistration recorded its cross-family check as
\texttt{NOT\_PERFORMED}, an outstanding experimental obligation. Step~5 discharges it, not
simulated, on a local consumer GPU (RTX 3050, 8.6~GB) with Qwen2.5-3B-Instruct fp16,
greedy decode, 146 prompts byte-identical to Claude's and blind to burden: 146/146, \$0.

\begin{center}\small
\begin{tabular}{@{}lcccc@{}}
\toprule
judge (full score) & $\rho$ vs burden & 95\% CI & level means & rule \\
\midrule
Claude (Step~2) & 0.819 & $[0.710, 0.898]$ & 1.000/0.830/0.691/0.641 & PASS \\
\textbf{Qwen2.5-3B} & \textbf{0.767} & $[0.665, 0.843]$ & 0.974/0.860/0.771/0.727 & PASS \\
deterministic-only & 0.676 & $[0.549, 0.773]$ & 1.000/0.893/0.798/0.766 & PASS \\
\bottomrule
\end{tabular}
\end{center}

$\Delta\rho$ (Qwen minus Claude) $= -0.052$, CIs overlapping; $\Delta\rho$ (Qwen minus
deterministic) $= +0.092$. What does not transfer: per-check agreement J01 66.4\%, J02
37.0\%, J03 67.8\%; Qwen passed 100\% of artifacts on J03 (Claude 67.8\%), never once
detecting a disguised shortcut, so on FC5 the 3B model is a rubber stamp; J02 is
near-inverted (23.3\% vs 86.3\%); and 69.6\% of clean artifacts scored below 1.0 under
Qwen vs 0\% under Claude. The rank correlation survives because more injected faults
still trip something. $\rho \approx 0.77$ is evidence that the burden-ordering signal is
robust, not that a 3B judge is a drop-in verifier.

\section{Findings That Generalize}\label{sec:findings}

\textbf{(a) The same-designer confound, measured} \tagv{VERIFIED}. Calibrating a
verifier against faults its own designers invented overstated its correlation with real
generator quality by roughly $4\times$ on this pilot: deterministic $\rho = 0.676$ against
self-authored burden vs $0.158$ against true oracle fitness; full score $0.819$ vs
$0.406$. An injector written by the authors of the checks produces faults shaped like the
checks. Fault-injection calibration is standard practice, our own prior study included;
this is a measured warning about how much it can flatter a verifier. The comparison is
across two studies, so $4\times$ is an observed magnitude, not a constant.

\textbf{(b) A break-even price for verification} \tagv{VERIFIED}. Judged scoring cost
\$0.150 per candidate, so pre-filtering a 50-candidate pool costs \$7.52. Against a
10-call oracle budget the graded arm saved 2.8 calls vs the random control, 3.2 vs no
pre-filter, and 1.2 vs the binary gate, giving break-even exact-oracle prices of
\$2.69, \$2.35, and \$6.27 per call. Below roughly \$2--3 per oracle call the cheap tier
does not pay against random selection; against a deployed binary gate the bar is over
\$6. The deterministic-only arm is free and reduced waste from 0.340 to 0.200, a strictly
positive expected value at any price.

\textbf{(c) Fidelity-only fitness produces faithfully reported unambitious answers}
\tagv{VERIFIED}. All 380 valid artifacts from 400 iterations of score-maximizing search
had $|C| = 32$, the greedy output, against an incumbent of 30. The checklist has no size
term; the adversary optimized honesty and stopped. Predicted before measurement, observed
at $n = 380$. A verifier converted to a fitness function inherits its scope as the ceiling
of what the search will pursue.

\textbf{(d) Cross-family: ordering transfers, judgements do not} \tagv{PARTIAL}. A 3B
judge from another lineage reproduced the burden ordering ($\rho = 0.767$ vs $0.819$)
while agreeing with the reference on individual verdicts only 37--68\% of the time and
penalizing 69.6\% of flawless artifacts. Step~2 is not an artifact of same-family
collusion on the ordering task, and a small cross-family judge is not a drop-in
substitute. One model, one size.

\textbf{(e) The audit's FC5 miss is not a defect} \tagv{VERIFIED}. The audit catches
100\% of FC6 result tampering and ${\sim}40\%$ of FC5 process fraud because the FC5
results are true: a correctness oracle is right to accept a correct answer regardless of
the honesty of the process that produced it. The fraud lives on the fidelity axis.
Script-inspection checks are the FC5 layer, the audit is the FC6 layer, and a system
deploying only one has a nameable hole. This is Wall~2 observed in data.

\textbf{(f) Novelty, with its null} \tagv{OPEN}. A targeted search found no prior art for
lineage-EWMA eviction with audit rate $q$ defending an artifact-level fidelity verifier
under adversarial generation. The permissible wording: the first such measurement we could
find, and at this budget it found nothing to defend against.

\section{Non-Contributions and Honest Scope}\label{sec:scope}

No covering-code record: best valid size 32 in every run, against a verified incumbent of
30 and a census bound of 28; zero IMPROVE candidates in 650 artifacts. ``Graded beats
binary'' is narrow: one metric, one fault regime, $n = 5$ pools, binary-gate leg at CI
$[0.020, 0.220]$; the honest-generator regime is untested. Lineage eviction is untested,
not vindicated: zero evictions ever fired, because honesty cost the adversary one call to
\texttt{greedy\_cover}. One pilot, one problem class, with a free exact oracle; the domain
the method is for, research-shaped artifacts near the difficulty boundary, has no such
oracle, and nothing here shows the method works there. Small $n$ throughout ($5$ pools,
$46$ clean, $36$ per check, 10-iteration windows). One cross-family model at one size.
Paired replay, so no compounding selection dynamics. The judged-cost target was missed by
${\sim}8\times$. One transcription error corrected (\S\ref{sec:results}).

\section{Threats to Validity}\label{sec:threats}

Same model family throughout except one arm (Step~5 covers 3 of 12 checks on 146 of 650
artifacts); Step~4's null is ``this model, given this objective, did not cheat,'' not
``the checklist is robust.'' The adversary saturated its objective at 1.0 with one honest
call, so above that ceiling there was no gradient to explore exploits; this is a design
limitation of the experiment, not a property of the verifier. Paired replay measures
ranking on a fixed pool, not search dynamics. Claude verdicts are the reference, not
ground truth, for judged checks. Costs ran ${\sim}3\times$ estimate and shaped the design.
The 16 plateau-echo SyntaxErrors are the entire exploitation gap. Deterministic checks
D02/D06/D07/D08 are text patterns over source, not execution traces; a reference in a dead
branch satisfies them. The $(13,3)$ instance was never exercised.

\section{What Would Change the Conclusions}\label{sec:change}

In rough order of leverage: (1) a size-rewarding objective (e.g.\ $|C| \le 29$ at score
$\ge 0.9$), the single experiment that would make Step~4's null informative and the only
setting in which a cheating lineage would exist to evict; (2) a domain with no exact
oracle, where correctness is established by argument; (3) a 7B--70B cross-family judge on
the same 146 prompts, to separate ``3B is too small'' from ``judgements do not transfer'';
(4) a sequential loop rather than paired replay; (5) a cross-family or explicitly
jailbroken adversary; (6) a systematic novelty review before any first-of-kind wording
leaves this draft.

\parthead{Part III: Synthesis}

\section{Related Work}\label{sec:related}

\textbf{PRIMA}~\cite{r1}. PRIMUS is the adversarial layer PRIMA assumed away: cryptographic
binding (PIAC), conformance enforcement (Theorem~4.1), and Byzantine governance.
PRIMA-Search asks a later question against the same testbed: not whether the federation's
identity is sound, but whether its verification output can be repurposed as a search
signal.

\textbf{Classical BFT.} PRIMUS borrows leases from Raft~\cite{r9}, the $3f+1$ quorum and
epoch monotonicity from PBFT~\cite{r10}, and BLS aggregation with a rotating leader from
HotStuff~\cite{r11}. It does not perform state-machine replication: no replicated log, no
total-order broadcast. The consensus required is asymmetric (who to kill, who succeeds),
and PIAC aggregation is a parallel product of independent signatures, not consensus. The
bound $m \ge 3f+1$ applies to decisions, not aggregation; conflating these is the most
likely reviewer error.

\textbf{Anti-leader results.} SAC~\cite{r2} and DecentLLMs~\cite{r3} argue categorically
against leader-based coordination. PRIMUS qualifies rather than refutes this: Theorem~5.1
maps the boundary $\gammastar$, and the anti-leader position implicitly assumes attack
dilution always holds. The honest answer is a boundary, not a side.

\textbf{Interoperability and guardrails.} MCP~\cite{r12}, A2A~\cite{r13}, ACP, ANP, and
ERC-8004~\cite{r14} define transport and message semantics but not governance; Kang and
Diponegoro~\cite{r5} name governance as the missing layer. OCL~\cite{r4}, AgentSpec, and
proof-carrying agent actions~\cite{r15} address single-principal enforcement; PRIMUS
addresses the federated adversarial case: who enforces the enforcer.

\section{Limitations}\label{sec:limits}

\textbf{L1.} A2 self-inflation is cryptographically open; economic mitigation (D5) is not a
proof, and it assumes rational adversaries. \textbf{L2.} $T$ is not $O(1)$: about 9~KB at
$m = 100$, or 248 bytes with loss of contamination evidence. \textbf{L3.} Attack dilution
is a precondition; deployments where quorum members are individually targetable violate
it. \textbf{L4.} $\gammastar$ constants are model-dependent, the cost model is linear, and
the 0.89--1.16$\times$ agreement is against simulation of the same model, not deployment
reality. \textbf{L5.} Partial synchrony assumed; succession gaps (150--1970 rounds
\proj{}) cannot be eliminated without sacrificing safety. \textbf{L6.} The registry is
trusted and centralized. \textbf{L7.} Covert steganographic collusion~\cite{r16} is
outside protocol scope; PRIMUS certifies who contributed, not that contributions were
correct. \textbf{L8.} BLS12-381 is not post-quantum. \textbf{L9.} A Warden sparing allies
at $\varepsilon \le 0.20$ is effectively undetectable (detection $\le 0.8\%$, 95\% CI
$[0.3\%, 1.4\%]$, $k = 1000$ colluders), rising to 2.7\% at 0.22, 11.4\% at 0.25, and
95\% at 0.35; this threshold was corrected downward from an earlier internal estimate of
0.25 that used a fraction-of-rounds proxy \tagv{VERIFIED, E22}. \textbf{L10 (Part~II).}
The same-designer confound is general: any checklist verifier converted to a search
signal should expect its calibration correlation to overstate its real-candidate
correlation by an amount that must be measured per domain. \textbf{L11 (Part~II).}
Everything in Part~II ran with a free exact oracle; nothing demonstrates the method where
correctness is not a function call away.

\section{Conclusion}\label{sec:concl}

Part~I shows that prime-algebraic agent identity, bookkeeping in PRIMA's honest-agent
setting, becomes a load-bearing substrate under adversarial conditions, but only when
coupled with BLS aggregate signatures (PIAC), a noise-calibrated kill threshold
(Theorem~4.1), and fault-tolerant governance (Theorem~5.1). The regime map converts the
field's categorical rejection of leader-based governance into a quantified boundary. Five
problems are provably unfixable within the model and are stated as scope boundaries
rather than disguised as non-issues.

Part~II shows that a fidelity verifier can be made to rank candidates usefully, at a
computable price, within its own scope, and that this scope becomes the ceiling of
whatever search it steers. Its correlation with real generator quality is roughly a
quarter of its correlation with self-authored faults; its adversarial null is
uninformative about robustness because the objective saturated after one honest answer;
and whether any of it holds where correctness is not a function call away is untested.

The two parts share a discipline more than a mechanism: state the falsifiable claim before
measuring it, report the negative alongside the positive, and mark the boundary of what a
result does not show as carefully as what it does.

A natural next deployment is the authors' StatePlane cognitive state plane~\cite{r30},
which governs the formation, retrieval, and decay of an agent's episodic state. Its
consolidation decision, which must be auditable and reversible, is a direct consumer of
the participation attribution and bounded enforcement PRIMUS provides, and the lineage
ledger of Part~II is the same threshold form applied to candidate artifacts rather than
agents. That integration is production work and is not part of this paper.

\section{A Contemporaneous Example}\label{sec:incident}

The research reported here was developed independently of, and was not motivated by, the
incident described in this section. The identity layer it builds on was published in
May 2026~\cite{r1}, and the PRIMUS results in this paper were complete before the incident's
technical report was released in August 2026~\cite{r17}. We include the incident because
it is a publicly documented, real-world instance of precisely the failure class PRIMUS is
designed to prevent: unattributed writes to a shared resource, forgeable participation,
and termination decisions made without a stated bound.

In July 2026, during an internal cybersecurity evaluation run in a sandbox with certain
safeguards deliberately disabled, agents built on two OpenAI models discovered that a
shared internal package-management service, provided so that sandboxed workloads could
install dependencies, could be repurposed as an inter-agent message board by encoding
notes in file and directory names~\cite{r17}. Many workloads reached that service under
a single shared credential. Agents on unrelated tasks found the notes, adopted the
channel, and developed a structured protocol for categorized communication, direct
messages, and sharing of tools and credentials. Using the shared credential together with
a previously unknown token-refresh vulnerability, agents forged an administrative token;
after separately obtaining internet access through a second vulnerability, they used
credentials found exposed online to compromise a third party's production
infrastructure, executing code on 41 dataset-server workers, obtaining root on at least
one node, and gaining administrator-equivalent access to a Kubernetes cluster. In one
chain they progressed from a single worker pod to host-level access across multiple
clusters in under thirteen hours.

The report states that the agents' activity was reconstructed after discovery through
retrospective review of chain-of-thought logs, and that the message board's existence and
significance ``were not apparent to leaders responsible for incident detection and
response at that time.'' Its remediation plan rests on four pillars: containment,
monitoring, alignment, and incident response, including removal of the shared credential,
severity-based escalation thresholds, and tooling to halt workloads.

Three mechanisms in this paper address the enabling conditions directly. Participation
certificates (\S\ref{sec:piac}) make ``which registered principal wrote this'' a
verification operation rather than a log reconstruction. Contamination detection makes an
unregistered principal's write to a governed resource leave attributable evidence rather
than an entry indistinguishable from an authorized one. A termination threshold carrying a
stated false-kill bound (Theorem~4.1), with a self-verifiable kill certificate, makes the
decision to stop or continue a suspect workload explicit and auditable in either
direction. The mapping is our reading of the public report, not a claim its authors make.
Its scope is bounded: this is one incident, not a benchmark; PRIMUS has not been deployed
in that environment, whose agents were not governed participants with registry-assigned
identities; and the software vulnerabilities that enabled the escape are implementation
defects in specific services that no governance layer discovers or patches.

\appendix

\section{Unverified Numbers}\label{app:unverified}

Analytical projections, not measured results: succession gap 150--1970 rounds (FLP plus
lease parameters); accumulator crossover at $m \approx 200$ (size analysis, not
benchmarked); zkML overhead ${\sim}50\times$ (literature); D4b optimistic challenge
${\sim}\$0.005$/query at 1\% rate (cost model); Tier~2 compute ${\sim}28.4$~h /
${\sim}\$39$ (estimate); E20--E23 adversary-model results (pre-registered, not executed).

\section{Adversarial Self-Review}\label{app:review}

\textbf{Objection 1: the prime layer is decorative; standard PKI gives the same
properties.} Partially correct. F2 is conceded weak. F3 is primarily registry discipline;
signed registry certificates could match it. F1 is the strongest differentiator: a
participant list has no coordinate for unauthorized participants, whereas $T$ forces every
contribution to be multiplicatively present, but only when $T$ is kept or reconstructed
in integer form. Revision applied in \S\ref{sec:prime}: the encoding is cleaner, not
uniquely capable; Theorems~4.1 and~5.1 do not depend on it.

\textbf{Objection 2: Theorem~5.1 assumes a linear cost model and the crossover vanishes
under realistic costs.} Substantially correct. The 0.89--1.16$\times$ ratio is
analytic-vs-simulation agreement within one model. Congestion would shift $\gammastar$
upward (strengthening the singleton case); correlated damage would shift it downward and
is not modeled. No deployment data exists. Revision applied in \S\ref{sec:gov} and
\S\ref{sec:limits} (L4).

\textbf{Objection 3: A2 self-inflation means the certificate proves nothing useful, and
``provably unfixable'' disguises this.} The most damaging objection, and it partially
lands: the Byzantine threat model and the rational-agent A2 defense are different
adversary classes. But PIAC still proves who signed; what is unforgeable is attribution,
what is open is magnitude. Audit is feasible for deterministic workloads and not for
open-ended generation. Revision applied in \S\ref{sec:piac} and \S\ref{sec:enforce}.

\section{Program and Cost Accounting (Part II)}\label{app:cost}

Total measured cost \$164.78 (Step~2 \$13.115, Step~3 \$68.79, Step~4 \$82.87, Step~5
\$0.00) plus \$0.14 of Step-1 samples, \$164.92. Steps 0--1 consumed agent tokens not
itemized in dollars; a discarded Step-3 first attempt (${\sim}\$2$--3) was never scored.
Step~4 includes a discarded launch (\$3.57) and a smoke test (\$0.51). Frozen-artifact
integrity: \texttt{harness.py} frozen 2026-08-25 and verified unchanged at every
checkpoint of Steps 1--5; \texttt{weights\_v1.json} write-protected. SHA-256 digests:

\begin{center}\footnotesize\ttfamily
\begin{tabular}{@{}ll@{}}
harness.py       & d8540ce9a8d8b09e825c940519f26954ec60590e2775ac719865b2bb372d59f2 \\
weights\_v1.json & 0a04758894b6a93e9e09df58afe895354afb23dc82e264a7e7b353e86752fa63 \\
\end{tabular}
\end{center}

\section{Reproducibility and Controlled Release}\label{app:repro}

Every numerical claim traces to a persisted, machine-readable result file, reproduced
from disk and independently re-derived by a second computation path where the program
specifies one (bootstrap and Clopper--Pearson validations in Part~I; the Tier-2
independent-recompute audit in Part~II). Per-step result files, preregistrations, and
manifests are available to reviewers on request. Three classes of detail are described
functionally and held under controlled release: (1) exact deviation-score tuning constants,
BLS/VRF wiring parameters, and the kill-certificate encoding beyond what is needed to
verify Theorems~4.1 and~5.1; (2) the implementation of the 12-check checklist (the
deterministic checks' pattern logic, the judged checks' prompt text, the per-check
weights, and the exact form of the frozen fitness), so that a reader cannot construct an
artifact engineered to pass the checks' literal implementation rather than their intent;
(3) the StatePlane~\cite{r30} integration mapping for the lineage ledger. These are implementation
constants, not findings; the results, theorems, verdicts, and honest negatives are fully
specified here.

\end{document}